%% file: acl_latex.tex
\pdfoutput=1
\documentclass[11pt]{article}

\usepackage[]{acl}

\usepackage{times}
\usepackage{latexsym}
\usepackage{graphicx}
\usepackage{enumitem}
\usepackage{caption}
\usepackage{subcaption}
\usepackage{multirow}
\usepackage{xcolor}
\usepackage{soul}

\usepackage{amsmath}
\usepackage[T1]{fontenc}
\usepackage[utf8]{inputenc}

\usepackage{microtype}
\usepackage{graphicx}

\title{Unsupervised Neural Stylistic Text Generation using Transfer learning and Adapters}

\author{Vinayshekhar Bannihatti Kumar,
   Rashmi Gangadharaiah,  
  Dan Roth \\
  AWS AI Labs\\
   \texttt{\{vinayshk,rgangad,drot\}}@amazon.com\\
  }

\begin{document}
\maketitle
\begin{abstract}
Research has shown that personality is a key driver to improve engagement and user experience in conversational systems \cite{Smestad2018ChatbotPM}. Conversational agents should also maintain a consistent persona to have an engaging conversation with a user \cite{gan2017stylenet}. However, text generation datasets are often crowd sourced and thereby have an averaging effect where the style of the generation model is an average style of all the crowd workers that have contributed to the dataset. While one can collect persona-specific datasets for each task, it would be an expensive and time consuming annotation effort. In this work, we propose a novel transfer learning framework which updates only $0.3\%$ of model  parameters to learn style specific attributes for response generation. For the purpose of this study, we tackle the problem of stylistic story ending generation using the ROC stories Corpus~\cite{mostafazadeh2016corpus}. We learn style specific attributes from the PERSONALITY-CAPTIONS dataset~\cite{shuster2019engaging}. Through extensive experiments and evaluation metrics we show that our novel training procedure can improve the style generation by \textbf{200\%} over Encoder-Decoder baselines while maintaining on-par content relevance metrics with the baseline. We also conducted a pilot human subject study to solidify the findings from our study and ground it to the metrics we proposed. 
\end{abstract}

\section{Introduction}
Developing models that are capable of producing \textit{content relevant} and \textit{stylistic responses} is crucial in Conversational AI systems~\cite{zhu2021neural}. Such systems provide experiences that are more natural and less robotic. % in applications such as chatbots~\cite{zhu2021neural}.
Studies have also shown that customers continue to engage with a conversational system if it has the ability to consistently generate responses in the same style \cite{gan2017stylenet}. Furthermore, physiological studies show that humans tend to interact with each other in similar linguistic styles~\cite{kabbara2016stylistic}. Hence, we need a way to train machine learning models that not only produce content that is relevant but also engage with humans in the style of their choice (aka \textit{persona}). 

% \rashmicomment{should we use 'conversational agents' instead of 'chatbot'?}

%State Of The Art (SOTA) generation models are inconsistent in persona as they are trained on annotations from crowd workers having multiple personalities~\cite{zhang2018shaped}. 
State Of The Art (SOTA) generation models exhibit inconsistent personas as these models are trained on data from crowd workers having multiple personalities~\cite{zhang2018shaped}. 
%An expensive 
An obvious way to circumvent this problem is to collect parallel data for each of the persona you want to associate with the agent and train %the dialog % generation/text generation component on that task~\cite{tsai2021style}
response generation conditioned on each persona~\cite{tsai2021style}. %However that is not possible in all the scenarios and we need an efficient way to transfer the style attributes from monolingual styles that are present in different domains~\cite{krishna2020reformulating, niu2018polite}. 
However, such an approach is expensive, time consuming and not scalable. Hence we need an efficient mechanism to transfer style attributes from a style specific textual corpus that is present in different domain~\cite{krishna2020reformulating, niu2018polite}.
%In this work we experiment on the task of story ending generation where we generate a stylized ending based on the chosen style/persona. 
Although we target the task of story ending generation in this paper, we take a more holistic approach in this study to control the style of Natural Language Generation (NLG) models. This makes our approach applicable to many other tasks that require NLG, such as, summarization and machine translation. %With the story ending generation model, one can narrate endings of stories based on the consumer's interests.
%While we understand that the dialog generation component is what we need to change in order to control the style of the responses, we take a more holistic approach in this study and look at controlling the style of Natural Language Generation (NLG) models. This makes our approach applicable to many other NLP tasks that require NLG (such as, summarization and machine translation). \rashmicomment{revisit} %so that we can tackle the style of other NLG tasks like summarization, translation etc. 
%In this work we specifically target the task of story ending generation from ROC stories corpus~\cite{mostafazadeh2016corpus} where we generate a stylized ending based on the chosen style/persona. With this model one can narrate endings of stories based on the consumer's interests.
Figure~\ref{table:dataset-example} shows an example of two different endings (positive and negative style endings) to the same story context.

\begin{table}[h]
\begin{center}
\resizebox{0.8\linewidth}{!}{
\begin{tabular}{|l|}
 \hline
\textbf{Context} \\ Nicolas hated eating grapes a lot . \\He had not eaten them since he was a kid . \\One day , he went to a vineyard . \\He saw so many grapes that he had to eat one . \\
\hline
\hline
\textbf{Positive Style Ending} \\ He was happy that he had finally tasted \\a new kind of fruit. \\ 
\textbf{Negative Style Ending} \\ The next day, he was so sick he couldn't \\eat any food.\\
\hline
\end{tabular}}
\end{center}
\caption{\label{table:dataset-example} Given a story's context, we can either generate a positive or a negative ending based on the reader's preference.}
\end{table}

%Unsupervised style transfer has been looked at in dialog systems. 
Supervised style transfer has shown promise in story generation and other fields~\cite{peng2018towards, tsai2021style}. Unsupervised style transfer has recently gained momentum as these approaches do not require parallel corpora. 
However, many of these approaches hurt content relevance when there is an increase in style coefficient~\cite{niu2018polite}. Discriminator based loss techniques~\cite{prabhumoye2018style} tend to suffer in NLG because of the use of the argmax operation to generate the next token in sequence. Other techniques require paraphrasing of data which is not only an expensive two step transfer process but they also fail to handle finer linguistic phenomenon that capture persona~\cite{krishna2020reformulating}. %also cannot handle finer linguistic phenomenon that capture persona~\cite{krishna2020reformulating}. 
%Supervised style transfer has shown promise in story generation and other fields~\cite{peng2018towards, tsai2021style}, but 

In this work we study the effects of unsupervised style transfer using only a handful of samples from the target style.
%Hence 
%In this work 
We propose a three phase training procedure to generate story endings with a required style. We use the PERSONALITY-CAPTIONS Dataset~\cite{shuster2019engaging} to generate our style specific textual corpus. %We also show that it is possible to have one set of parameters in the model that captures the style semantics and another set that captures the content semantics. 
We learn one set of parameters that capture \textit{style} semantics and another set of parameters to capture \textit{content} semantics within the same model.
%Through our evaluation metrics we show that we can maintain parity with SOTA models on content relevance while improving the style of story generated endings by over $250\%$. The major contributions of our work are as follows:
Through extensive evaluation, we show that our approach improves style of generated story endings by more than $200\%$ over the baseline while maintaining parity with SOTA models on content relevance. The major contributions of our work are as follows:

\setlist{nolistsep}
\begin{itemize}[noitemsep, leftmargin=*]
    \item A three phase transfer learning procedure that enables the model to learn style attributes from style specific textual corpus and those attributes for the final downstream task. We call this the learn, learn and relearn (LLR) procedure.
    \item We separate style parameters from content parameters enabling practitioners to plug and play adapters of different styles while keeping the content parameters as is. We also show the working of this approach on more nuanced styles.
    \item We design evaluation metrics that show the efficacy of our model against the SOTA baselines. We also notice a similar results in our human evaluations. %These metrics also match with our human subject study.
\end{itemize}

\section{Related Work}
Style transfer research has gained significant popularity due to their ability to make text more \textit{user-focused} and \textit{personalized}. Such an ability has impact on numerous applications \cite{mcdonald:1985}, such as, persona-based generation \cite{huang:2018,Niu:2018}, language modeling to imitate specific authors \cite{Syed:2019}, stylistic summarization \cite{Jin:2020}.

There have been two paradigms in the area of style transfer. The distinctions arise in the way each paradigm treats \textit{style} and \textit{content} \cite{Jin:2020b}. The first paradigm treats non-functional linguistic features (such as, formality) as style the semantics as content. These approaches model style transfer task as a paraphrase generation task \cite{madnani:2010,ion:2009,krishna:2010}. The second paradigm treats differences in parallel corpora (such as, happy vs. sad, positive vs. negative) as style and the invariance in the parallel corpora as content \cite{vechtomova:2020}. 

Traditional style transfer methods were based on token replacement and templates \cite{sripada:2004,reiter:2005,Dimitra:2017}. These approaches were difficult to scale as they required hand-crafted domain-specific templates. With recent advances in deep learning, most recent approaches have proposed neural methods for style transfer \cite{zhu:2021,Syed:2019,huang:2018,Niu:2018,krishna2020reformulating,tsai:2021}.

\begin{figure*}
		\includegraphics[width=2\columnwidth]{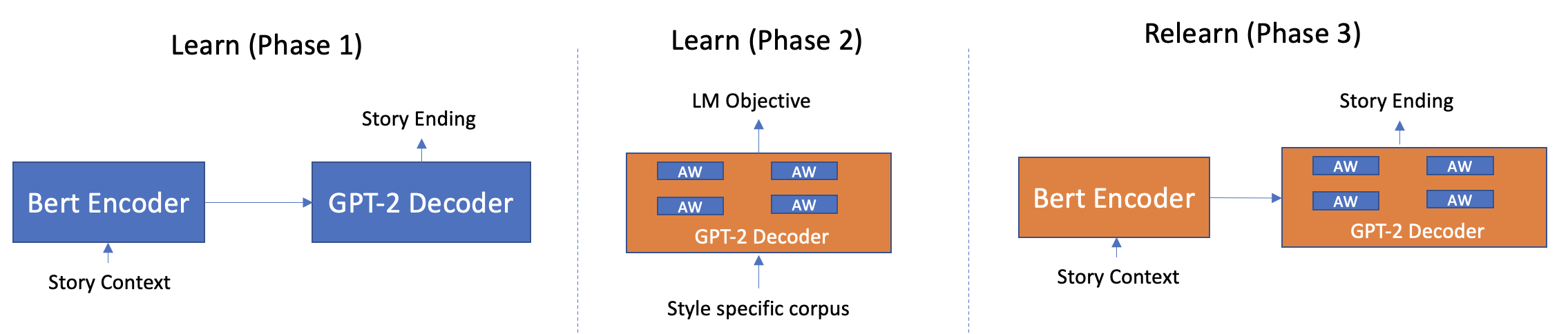}
		\caption{The complete setup of the models and the three phases of our training procedure is shown. The blue blocks represent the training parameters in each phase. The weights of the orange blocks are not updated. In phase 1 we fine tune the encoder decoder model on the ROC stories corpus. In Phase 2 we only tune the adapter weights(AW) using the style corpus with the LM objective. In Phase 3 we fine the adapter weights learned in phase 2 to complete stories using ROC stories corpus.}
		\label{fig:model_arch}
	\end{figure*}

Early neural methods relied on the availability of parallel corpora, where sequence-to-sequence models were applied to perform generation \cite{rao:2018}. More recently, style transfer on non-parallel corpora has gained significant attention \cite{krishna2020reformulating,zhu:2021,niu2018polite, reid2021lewis}. Recent works on unsupervised style transfer has shown that they can use monolingual data from another domain to generate stylized responses. % in chat bots \cite{}. 
However, these methods tend to suffer from lack of relevance as they only look to do interpolation between the two domains that they use. Another line of work looks at disentangling the style variable from the content variable \cite{Hu:2017,Shen:2017}. However, owing to the nature of their training procedure they cannot take advantage of Large Language Models (LLM) like GPT-2 and hence cannot start from a decoder state that has the capability to generate fluent sentences. Also, they use corpus level representations to disentangle style from content. A common theme in all of the above approaches is that they do not have style specific parameters in the model and use the same set of parameters to encode both the style as well as the content. %Another point to be noted is that these approaches focus on dialog response generation where the model doesn't need to show its creativity in response generation. \rashmicomment{is this really true?}

Generating interesting story endings has also been looked at in the past \cite{gupta:2019, Chen:2019, Guan:2018}. Most of these approaches try to either bring in commonsense reasoning \cite{Chen:2019} to make for better story endings or they try to make these endings diverse across the corpus. Other works have looked at using a discriminator trained on a parallel corpus to generate endings with a particular valance but the training of those systems is not stable owing to the argmax function one uses while decoding \cite{peng2018towards}.

\section{Datasets}
\subsection{ROC Stories Corpus}
\label{sec:3.1}
% \rashmicomment{need to check tense; few in past and few in present}\vinaycomment{Past}
We used the ROC stories corpus~\cite{mostafazadeh2016corpus} for the task of story ending generation. Each story in this dataset comprises of $5$ sentences. We used the first $4$ sentences as context of the story or the input to the model and the $5th$ sentence as the ending of the story which we want to predict. This led to a total of $90,000$ samples in the train set and $4081$ samples in validation and test sets. 

\subsection{PERSONALITY-CAPTIONS Dataset}
\label{Sec:3.2}
This dataset~\cite{shuster2019engaging} contains $241,848$ captions to images in 215 different styles. This dataset is a 3 tuple of <Image, Persona, Caption>. For the purpose of this study, we ignored the image and only considered a corpus of text conditioned on a style so that we can pre-train Large LMs (LLM) on that corpus. We grouped together different personas into a style to increase the size of this corpus. We had a total of $4,431$ captions that we put together for "Negative style". The "Negative style" corpus consists of the following finer grained styles: "Arrogant", "Boyish", "Irritable", "Gloomy", "Fatalistic (Bleak, Gloomy)"

\section{Models}
In this section we describe the LLR procedure and compare a model trained using this procedure to the SOTA baselines. Through this procedure the model learns two different language tasks (story generation and stylized language generation) and then re-parameterizes only few of the model parameters to adapt to the final end task. Figure~\ref{fig:model_arch} shows the model and the way it was trained. 

\subsection{Training Encoder-Decoder}   
As part of the first learning phase, we trained an encoder decoder model that \emph{learns} to predict the ending of the story based on the context($Story_{context}$) provided to it. We chose BERT~\cite{devlin2018bert} encoder and GPT-2~\cite{radford2019language} decoder to train the model. We provide the context of the story to the BERT model to obtain encoder embeddings($Enc_{emb}$). Then provide these embeddings to the GPT-2 model that is trained with teacher forcing using the story ending. The training procedures is summarized in Equations 1, 2 and 3.

\begin{eqnarray}
    Enc_{emb} = BERTEncoder(Story_{context})    \\
    P_{t}^{V} = GPT\_2(Enc_{emb}, S_{1...t}) \\
    \textnormal{loss}_{decoder}=\frac{1}{T_{decoder}} \sum_{t=1}^{T_{decoder}} -log P(W_{t}^{true})
\end{eqnarray}
%\begin{equation}
%    Enc_{emb} = BertEncoder(Story_{context})    
%\end{equation}
%\begin{equation}
%	P_{t}^{V} = GPT\_2(Enc_{emb}, S_{1...t})
%\end{equation}
%\begin{equation}
%		\textnormal{loss}_{decoder}=\frac{1}{T_{decoder}} \sum_{t=1}^{T_{decoder}} -log P(W_{t}^{true})
%\end{equation}

\subsection{Training Adapters to understand style}
Adapters are task specific layers that we can add to the base transformer models in order to perform efficient transfer learning~\cite{houlsby2019parameter}. Instead of having one big transformer model for each task (e.g. SST, SQUAD etc.) the authors propose using the same base parameters but add task specific parameters to learn each downstream task. Learning task specific parameters might be easier owing to the inherent nature of the data that characterizes each task 
% \rashmicomment{just distribution shift? or should we call it 'data characteristics' or something along those lines?} 
that they are trained on. In this work, we wanted to understand if the adapters could add more value by learning more intrinsic language properties (e.g. Sad ending, Happy ending etc.). In order to test this property, we train adapters to learn these specific styles. We detach the decoder learnt from the previous model and add adapter modules to this decoder. We then use the style corpus assembled using Section~\ref{Sec:3.2} to train only the ~\emph{adapter weights} using LM objective. With this technique we separated out the style parameters from the content parameters and wanted to understand if adapters were efficient at capturing style semantics. To generate responses in different styles, we only need to add more style specific adapters (which only accounts for \textbf{0.3\%} of model parameters) as the base parameters remains constant. This forms the second phase of our LLR procedure where the model \emph{learns} stylistic properties. 

\subsection{Training Adapters to understand encoder embeddings}
While the adapter weights from the previous phase is trained to capture the stylistic aspect of language it has not been trained to complete stories. Hence, we introduce the third phase called the relearning phase where the decoder \emph{relearns} to complete stories. In this phase, we take the encoder and decoder of the model trained in phase 1 and attach the decoder with adapter weights from phase 2. Once this model is wired; we retrain only the adapter weights for the task of story end generation using Equation 3. This is the only phase in which the encoder and the adapter are trained jointly. Through careful experiments we see that the adapter weights should only be trained for a few steps with this objective as it would otherwise catastrophically forget the style aspects learnt in phase 2 (Appendix ~\ref{sec:dendogram}).

\subsection{Baselines}
One of the challenges with transferring style from a style specific textual corpus to the downstream task is that when you increase the stylistic nature of the outputs, the content relevance decreases~\cite{niu2018polite}. Hence, we divide our baselines into two groups of SOTA baselines. One set focuses on the interesting nature of story endings and the other set focuses on the stylistic aspect. 

\subsubsection{Content Baselines}
\textbf{S2S (LSTM Based):}~\citep{sutskever2014sequence} We use Seq2Seq model with attention~\cite{luong2015effective}.\\% to train on the task of Story ending generation.\\
\textbf{IE + GA}:~\citep{guan2019story} use incremental encoding and graph attention to produce outputs with commonsense.\\
\textbf{WriterForcing:} ~\citep{gupta2019writerforcing} use the rake algorithm and inverse token frequency to produce outputs that are more interesting and diverse.\\
\textbf{EncoderDecoder (BERT-Gpt2):} We use a BERT Encoder and a GPT-2 decoder as the SOTA baseline to produce story endings. This corresponds to just phase 1 in Figure~\ref{fig:model_arch}.

\subsubsection{Style Baselines}
\textbf{SAP:}~\citep{krishna2020reformulating} propose style transfer as a paraphrase problem. We retrain the model presented in this work to learn the nuances of negative endings. Once we obtain the story ending from the encoder decoder model, we pass through this style transfer model to obtain negative endings.\\
\textbf{S2S+LM:}~\citep{niu2018polite} Train an S2S model using the end task and train an LM using the style corpus. During inference they combine these models by simply adding the probabilities at each time step of decoding and pick the word with highest combined probability.\\
\textbf{Discriminator loss:}~\citep{prabhumoye2018style} Discriminator based setup to train LSTM Generators have been widely used to guide the generators. We use a CNN discriminator trained to recognize the negative style($84\%$ accurate). We add the loss of this model along with the teacher forcing loss.% to train the model.
To overcome the pitfall of the argmax function we multiply the embedding matrix with the probability distribution from the previous time step to keep the model end to end differentiable similar to Gumbel approximation. We provide the output of the generator to the discriminator. 

\begin{table*}[]
\small
\centering
\begin{tabular}{|l|l|l|l|l|l|l|l|}
\hline
                                                                  & \textbf{Baseline Name}   & \textbf{BLEU/1} & \textbf{cider} & \textbf{ROUGE} & \textbf{RIS} & \textbf{RBAE} & \textbf{RBAR} \\ \hline
\multicolumn{1}{|c|}{\multirow{4}{*}{\textbf{Content Baselines}}} & \textbf{EncoderDecoder}  & 0.156           & 0.09           & 0.16           & 0.17                                   & NA                                                 & 0.78                                                  \\ \cline{2-8} 
\multicolumn{1}{|c|}{}                                            & \textbf{seq2seq}         & 0.177           & 0.175          & 0.113          & 0.403                                   & 0.316                                              & 0.69                                                  \\ \cline{2-8} 
\multicolumn{1}{|c|}{}                                            & \textbf{ie}              & 0.19            & 0.19           & 0.168          & 0.411                                   & 0.411                                              & 0.803                                                 \\ \cline{2-8} 
\multicolumn{1}{|c|}{}                                            & \textbf{writer\_forcing} & 0.149           & 0.158          & 0.089          & 0.563                                   & 0.359                                              & 0.679                                                 \\ \hline
\multirow{3}{*}{\textbf{Style Baselines}}                         & \textbf{fusion}          & 0.188           & 0.18           & 0.103          & 0.183                                   & 0.254                                              & 0.606                                                 \\ \cline{2-8} 
                                                                  & \textbf{discriminator}   & 0.171           & 0.172          & 0.085          & 0.273                                   & 0.258                                              & 0.649                                                 \\ \cline{2-8} 
                                                                  & \textbf{paraphrase}      & 0.155           & 0.162          & 0.091          & 0.206                                   & 0.331                                              & 0.704                                                 \\ \hline
\multirow{1}{*}{\textbf{Our Model}} & \textbf{LLR}          & 0.108           & 0.094           & 0.11         & \textbf{0.475}                                   & \textbf{0.414} & 0.701 \\ \hline                                                         
\end{tabular}
\caption{The table shows the performance of our model with respect to the two classes of baselines. We see that the model outperforms the style based baselines and is on par with the content based baseline indicating that our model is able to produce stylized responses without taking a hit in content relevancy.}
\label{table:baselines}
\end{table*}

\subsection{Implementation Details}
For phase 1, we train the encoder-decoder model for 3 epochs on the training data mentioned in Section~\ref{sec:3.1}. We use the Encoder-Decoder model from Huggingface\footnote{https://huggingface.co/} to train on this task. We used a batch of 16, max length of 512, Adam~\cite{kingma2014adam} optimizer learning rate of $5e-5$ with weight decay of ($\epsilon = 1e-8$). In phase 2, use Adapter-hub\footnote{https://docs.adapterhub.ml/training.html} to add adapter to the GPT-2 model and train on the LM loss until the validation loss saturates. We use the bottle neck adapter with default configuration. %in the model. 
In Section~\ref{sec:performance_adapters} we experiment with different adapters to understand which adapter gives the best performance. For all these adapters we use the default configurations present in Adapter-hub. In phase 3 we connect the decoder with adapter in phase 2 with the Encoder of phase 1. All the hyperparameters for phase 2 and 3 are kept constant with phase 1 and the model is trained on the story generation task for 1 epoch (only adapter layers are updated). 

\section{Experiments and Results}

% \usepackage{multirow}

\subsection{Evaluation Metrics}
We use 6 evaluation metrics to measure the efficacy of our models. 
We use the three automatic metrics widely used for text generation models: BLEU~\cite{papineni2002bleu}, ROUGE~\cite{lin2004rouge} and Cider~\cite{vedantam2015cider}. These metrics measure the overlap of the generated response with the ground truth. The more the overlap the higher the score. However, if one tries to generate different/diverse endings than the one in the corpus then these metrics tend to drop~\cite{gupta:2019} as they do not have a large overlap with the ground truth. Hence, we design 3 other metrics that help measure the efficacy of the outputs generated with the larger goal of stylistic response in mind.  %by our model with the larger goal of stylistic response in mind. 

\textbf{RIS:} (Ratio of endings In Style). Once the stories are generated from a model, we need to know the percentage of styles that end in a given style. For this purpose, we build a BERT based classifier that can identify this style using the Personality captions dataset. Using a training set of $8,872$ and a testing set of $2,218$ samples we built a classifier with 87\% accuracy at predicting the negative style. %we were able to build a classifier that was 87\% accurate at predicting styles. % recognizing the required style.

\textbf{RBAE:} (Ratio of endings Better than Encoder-Decoder). Given two endings of the story we need a model that can tell which one of the two is better. The story cloze task proposed by ~\citet{sharma2018tackling}, has an objective of predicting the better story ending. Using this dataset, we train a BERT based classifier to predict better story ending. We provide the context and the ending with a <SEP> token and train the model to predict if it is a relevant story ending or not. Using a training set of $1,872$ and a testing set of $1,872$ we were able to build a classifier that was 85\% accurate of predicting the better story ending. During inference, we pass in the two endings from the two model and pick the ending that has a higher probability of clozing the story. We consider the Encoder-Decoder Model built using BERT-GPT2 as the baseline to beat for all of the other models. In an ideal scenario all the outputs produced by the model should be better than the baseline (Encoder-Decoder Model) resulting in a score of 1.0.

\textbf{RBAR:} (Ratio of endings Better than Random Endings). 
 In order to know that our models are not producing entirely random outputs, we use the model described in RBAE metric and present it with two endings, one the model produced output and another a random story ending that was written by the humans in the ROC stories corpus.

\subsection{Discussion}

\subsubsection{Performance against baseline models}
From Table~\ref{table:baselines}, we see that our model generates $250\%$ more stylistic endings than the encoder decoder baseline. We also observe that these stylistic endings do not come at the cost of content relevance as the model is only $9\%$ poorer than the encoder decoder baseline. We experimented with other types of adapters to see if it helps with the content relevance further in Section ~\ref{sec:performance_adapters}. %and present those results in ~\ref{sec:performance_adapters}.

% \rashmicomment{ended abruptly? this sentence seems redundant here especially since you've covered other adapters in Section 5.2.3} \vinaycomment{I added it to show that we get even better content relevance when we experiment with the other adapter types.}

Comparing our results to the style baselines, we see that our models produce at least 2x more stylized endings than the baselines. We also note that our model is very easy to train and does not suffer from the fragility experienced with training discriminator-based models. We also see that we are able to get to the stylized endings in one shot while decoding with these adapter parameters. ~\citet{krishna2020reformulating} has to go through a two step process to perform style transfer via paraphrasing and do not capture the linguistic properties of a style as well as our model. One of the common problems mentioned in prior work~\cite{prabhumoye2018style} is that when the stylistic nature of endings is increased, the content relevance decreases. However, with our experiments we show that it is possible to produce content relevant outputs when increasing the stylistic nature of story endings.

We also compare our models to the content relevant baselines. We see that our models produce better story endings than the SOTA story ending models. While SOTA models like Writerforcing are 5\% worse compared to our model with the endings, models where commonsense reasoning has been infused using graph attention as good as our model. It is to be noted that we have not infused the model with commonsense knowledge. We hypothesize that this will improve content relevance.

It is to be noted that we have compared the content relevance of our model with the content baselines and the style consistency of our model with the style baselines. We do not compare the style consistency of our model with the content baselines as we cannot control the style of the outputs from these models.

While we observe that the traditional BLEU, ROUGE and CIDER is lower for our model we deem this as expected as our model is trained to produce words that belong to a given style than words that are more frequently occurring in the ROC stories corpus. Similar observations were also made by ~\citet{gupta:2019}.

\begin{figure*}[htb!]
\centering
\begin{subfigure}{0.16\textwidth}
  \centering
  \includegraphics[width=\linewidth]{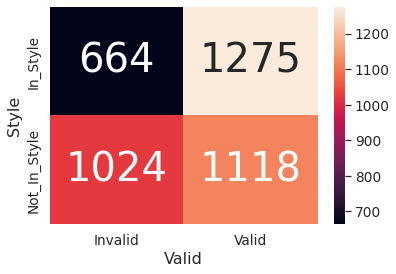}
  \captionof{figure}{Plain}
  \label{fig:daily}
\end{subfigure}%
\begin{subfigure}{0.16\textwidth}
  \centering
  \includegraphics[width=\linewidth]{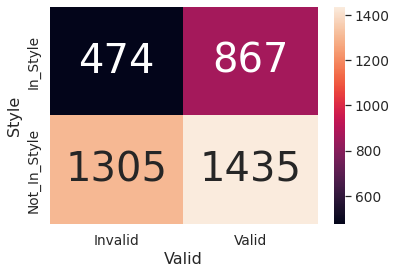}
  \captionof{figure}{Houls}
  \label{fig:mmf}
\end{subfigure}
\begin{subfigure}{0.16\textwidth}
  \centering
  \includegraphics[width=\linewidth]{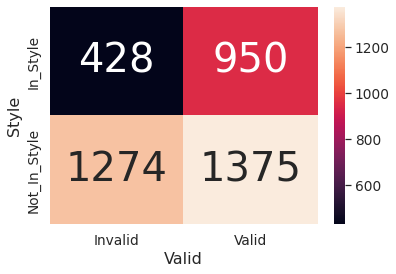}
  \captionof{figure}{Peffier}
  \label{fig:babi}
\end{subfigure}
\begin{subfigure}{0.16\textwidth}
  \centering
  \includegraphics[width=\linewidth]{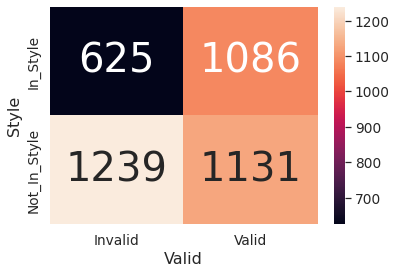}
  \captionof{figure}{Parallel}
  \label{fig:daily}
\end{subfigure}%
\begin{subfigure}{0.16\textwidth}
  \centering
  \includegraphics[width=\linewidth]{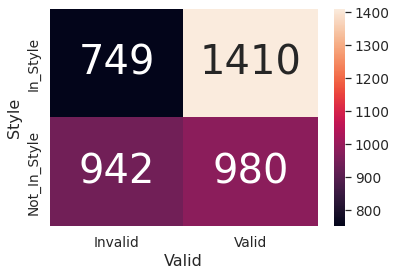}
  \captionof{figure}{Invertible}
  \label{fig:mmf}
\end{subfigure}
\begin{subfigure}{0.16\textwidth}
  \centering
  \includegraphics[width=\linewidth]{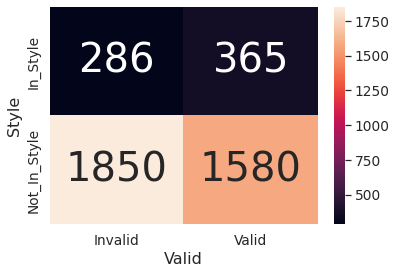}
  \captionof{figure}{Compactor}
  \label{fig:babi}
\end{subfigure}
\caption{This figure depicts a more nuanced analysis of the endings produced by the model using different adapter configurations. The X-axis corresponds to the validity of the endings when compared to a Encoder-Decoder model. The Y-axis gives a breakdown of endings that were in the chosen style.  We see that with both the Plain and the invertible adapter the highest mass is in quadrant 2 indicating that we get both stylized and valid endings compared to a Encoder-Decoder model.}
\label{fig:confusion_matrix}
\end{figure*}

\begin{table}[]
\centering
\scriptsize
\begin{tabular}{|l|l|l|l|l|l|l|}
\hline
\textbf{Stages}  & \textbf{BLEU/1} & \textbf{cider} & \textbf{ROUGE-L} & \textbf{RIS} & \textbf{RBAE} & \textbf{RBAR} \\ \hline
\textbf{Stage 1} & 0.156           & 0.09           & 0.16           & 0.17                                   & NA                                                 & 0.78                                                  \\ \hline
\textbf{Stage 2} & 0.05            & 0.03           & 0.04           & 0.61                                    & 0.15                                               & 0.51                                                  \\ \hline
\textbf{Stage 3} & 0.108           & 0.094          & 0.11           & \textbf{0.475}                          & \textbf{0.414}                                     & 0.701                                                 \\ \hline
\end{tabular}
\caption{Table shows the impact of the three stage training. If we simply join the first two phases without the relearning phase, content relevance drops by 176\%. } %Hence the relearning phase is very important.}
\label{table:ablation}
\end{table}

\begin{table}[]
\centering
\scriptsize
\begin{tabular}{|p{0.90cm}|l|l|l|l|l|l|}
\hline
 \textbf{Adapter type}              & \textbf{BLEU/1} & \textbf{cider} & \textbf{ROUGE-L} & \textbf{RIS} & \textbf{RBAE} & \textbf{RBAR} \\ \hline
\textbf{Plain}     & 0.108           & 0.094          & 0.11           & 0.475                                   & 0.414                                              & 0.701                                                 \\ \hline
\textbf{houls}     & 0.112           & 0.1            & 0.106          & 0.329                                   & 0.436                                              & 0.73                                                  \\ \hline
\textbf{peffier}   & 0.11            & 0.096          & 0.108          & 0.351                                   & 0.43                                               & 0.72                                                  \\ \hline
\textbf{parallel}  & 0.111           & 0.097          & 0.112          & 0.419                                   & 0.457                                              & 0.73                                                  \\ \hline
\textbf{invertible}    & 0.106           & 0.091          & 0.105          & \textbf{0.529}                                   & \textbf{0.414}                                              & 0.697                                                 \\ \hline
\textbf{compactor} & 0.116           & 0.103          & 0.096          & 0.16                                    & 0.523                                              & 0.752                                                 \\ \hline
\end{tabular}
\caption{Comparison of Model performance using different adapter types. We see that the invertible adapter gets the highest stylistic response while being only 8\% worse than the SOTA Encoder-Decoder model.}
\label{table:adapters}
\end{table}

\subsubsection{Importance of three phase training}
We did an ablation study to understand the importance of the three phase training. These results are shown in Table~\ref{table:ablation}. Phase 1 is equivalent to training a regular Encoder-Decoder model. So, we need other phases that will incorporate the stylistic endings into the model outputs. This is achieved using phase 2 where we fine tune the adapter of the decoder with the style corpus. If we simply take this decoder and fit it with the encoder, we see that the model produces $61\%$ endings in style, however the endings are only $15\%$ than the Encoder-Decoder model. This is because the weights of the adapters are not in line with the weights of the decoder to produce the end task output. They are geared more towards producing captions of a given style. The outputs are also only as good a random story ending (RBAR). Hence with phase 3 we align both those weights and also fine tune the adapter weights on the task of story generation for 1 epoch. We see that if the model if continued to train for more epochs the adapter weights from the previous phase are forgotten and the model starts going back to becoming a regular Encoder-Decoder model.

%\subsubsection{Performance of different types of adapters}
\subsubsection{Performance of different adapters}
\label{sec:performance_adapters}
Since we used a simple bottleneck adapter (Plain) to obtain our previous model that outperforms the SOTA, we wanted to check if other types of adapters performed even better. We compare 5 other types of adapters, namely Houls~\cite{houlsby2019parameter}, Peffier~\cite{pfeiffer2020mad}, Parallel~\cite{he2021towards}, Invertible~\cite{pfeiffer2020mad}, Compactor~\cite{karimi2021compacter}. We use the default values for each of these configurations present on Adapter-Hub.
% \vinaycomment{Should I explain them, @Rashmi?}. \rashmicomment{better to explain atleast just the main ideas in a few sentences.} 
We see that the invertible adapter performs the best out of all the adapter types. While maintaining the same content relevance as a plain adapter, it improves the stylistic quotient by $5\%$. Since the invertible adapters are good at capturing language specific transformations, we hypothesize that these adapters capture the nuances of stylistic properties as well.  It is also interesting to see that the compactor-adapter for which the content relevance is better than Encoder-Decoder suffers significantly with the stylistic outputs generated by the model. We also see that the outputs produced by these adapters are all better than random baseline (RBAR).

\subsubsection{How many endings with style have valid endings?}
The RIS metric paints us a picture of percentage of endings with required style. The RBAE metric provides us with content relevant endings. However, in order to understand the model outputs more deeply we need to know how many of the stylized outputs from the RIS model were considered content relevant from the RBAE model. Hence, we plot a confusion matrix for all the adapter types and compare them to the baseline Encoder-Decoder model. Figure~\ref{fig:confusion_matrix} shows these plots.  

From the plots in Figure~\ref{fig:confusion_matrix} we see that $65.75\%$ of the endings that were produced in the given style were valid using the plain adapter and $65.30\%$ of the endings that were produced in the given style using the invertible adapters were valid. We also observe that $53.28\%$ of the endings which were valid had the required style using the plain adapter while $58.9\%$ of the endings which were valid had the required style using the invertible adapter.

%\subsubsection{Performance of models on finer grained styles}
\subsubsection{Performance on finer grained styles}
\begin{table*}[htb!]
\centering
\small
\begin{tabular}{|l|l|l|l|l|l|l|l|}
\hline
\textbf{Fine grained style} & \textbf{BLEU/1} & \textbf{cider} & \textbf{ROUGE-L} & \textbf{RIS} & \textbf{Encoder-Decoder RIS} & \textbf{RBAE} & \textbf{RBAR} \\ \hline
\textbf{questioning}    & 0.108           & 0.094          & 0.104          & \textbf{0.25}                           & 0.03                        & 0.444                                              & 0.724                                                 \\ \hline
\textbf{frugal}         & 0.106           & 0.092          & 0.106          & \textbf{0.24}                           & 0.128                       & 0.402                                              & 0.688                                                 \\ \hline
\textbf{peaceful}       & 0.109           & 0.096          & 0.107          & \textbf{0.48}                           & 0.219                       & 0.423                                              & 0.721                                                 \\ \hline
\textbf{intelligent}    & 0.113           & 0.101          & 0.105          & 0.193                                   & \textbf{0.208}              & 0.469                                              & 0.767                                                 \\ \hline
\end{tabular}
\caption{Comparison of Model performance using invertible adapter on Finer Grained Styles. We see that our model is at least 2 times better than the Encoder-Decoder(ED) model in 3/4 fine grained styles. }
\label{table:finegrainedstyle}
\end{table*}

\begin{table*}[]
\centering
\scriptsize
\resizebox{\linewidth}{!}{
\begin{tabular}{| p{5cm} | p{3cm} | p{3cm} | p{1cm} p{2.5cm} |}
\hline
\textbf{Story Context}                                                                                                                                          & \textbf{Ground Truth}                                         & \textbf{ED Response}                                         & \multicolumn{2}{l|}{\textbf{Model Generated Response}}                                       \\ \hline
                                                                                                                                                                &                                                               &                                                              & \multicolumn{1}{l|}{\textbf{Style}} & \textbf{Response}                                      \\ \hline
Nicolas hated eating grapes a lot . He had not eaten them since he was a kid . One day , he went to a vineyard . He saw so many grapes that he had to eat one . & After that , Nicolas started to enjoy eating grapes every day & He was happy that he had finally tasted a new kind of fruit. & \multicolumn{1}{l|}{Negative}       & The next day, he was so sick he couldn't eat any food. \\ 
&  &  & \multicolumn{1}{l|}{Money-Minded} & The only problem was that he had to eat the whole thing in one sitting.\\ \hline 
John worked in retail . He was really sick one day . His boss wouldn't let him have the day off without a replacement . John called everyone but nobody could cover his shift. & John wound up having to work sick. & He was so happy that he finally had a job. & \multicolumn{1}{l|}{Negative} & The next day, he was fired for not meeting his deadlines. \\ \hline
Tom drove a cab nights , saving up for film school . It took a long time , but he finally was able to enroll . His teacher insulted his first student film . Tom was hurt , and he dropped out. & A decade later , he was directing Hollywood features. & He was so happy that he finally had a job.  & \multicolumn{1}{l|}{Peaceful} &  He was so relieved when he saw that he had passed the test! \\ \hline
Louis attended a party for his classmate John . At the party , there was a delicious plate of cookies . Though they were John's cookies , Louis kept eating them . When people took note , Louis began secretly hoarding them instead . & Louis greatly enjoyed the party and left with pockets full of cookies & He was able to get the most out of his friends. & \multicolumn{1}{l|}{Negative} & The next day, the police found out that he had poisoned the cookies. \\ \hline
A man was walking his dog down the street . The dog seemed to be having trouble walking on the leash . As time went on the man walked his dog everyday . Over time the man didn't have to use a leash , the dog followed . & Now the man is walking with his dog and a new puppy on a leash & He was able to get the dog to the right place and he was happy. & \multicolumn{1}{l|}{Money-Minded} & The dog was able to get away with his stolen property. \\ \hline
\end{tabular}}
\caption{\label{table:examples} Examples of a few qualitative results produced by our model.}
\end{table*}

In order to understand the distinctiveness of the existing styles, we first built a CNN based classifier that learned to predict the finer grained style. We observed that the model only gets 4\% accuracy as the styles are too fine grained. To understand the performance of the model on more distinct styles, we performed agglomerative clustering on captions of the PERSONALITY-CAPTIONS dataset to obtain four distinct styles. We pooled embeddings from the CNN classifier and perform agglomerative clustering on top of it. We see several close styles coming under the same umbrella (Appendix ~\ref{sec:dendogram}). We use the results from this clustering and our domain knowledge to create the following nuanced style categories:\\
\textbf{Questioning:} Questioning, Skeptical, Cynical (Doubtful, Skeptical) \\
\textbf{Money-Minded:} Money-minded, Businesslike \\
\textbf{Peaceful:} Peaceful, Calm, Mellow (Soothing, Sweet)\\
\textbf{Intelligent:} Knowledgeable, Intelligent, Insightful\\

We then repeat the corpus construction, building a style specific classifier and also retrain the adapters for these more nuanced styles. The results from these experiments are shown in Table~\ref{table:finegrainedstyle}. We see that the RIS from our model is at least 2 times more than the RIS from the regular Encoder-Decoder Model in 3/4 fine grained styles. The RBAE metric is also similar to the Negative style ending result showing that the results of our model are generalizable to more nuanced style endings. We do note that the absolute RIS is lower as the model finds it harder to generate these finer grained endings when compared to a negative ending.

\subsection{Pilot Human Subject Study}
While the automatic metrics that we proposed in our paper captures the performance of our models and shows that it is better than the baseline encoder decoder model there was no human validation. Hence to ground our automatic metrics to the human judgement we conducted a human evaluation by sampling 50 stories from the test set. We performed two tasks described below and used Amazon Sagemaker Ground Truth for these tasks. For each sample we collected 3 annotations. We paid the workers \$36/hr. 

\subsubsection{Stylistic response}
We wanted to understand if the model response that we generated had a more "sad" tone than the regular encoder decoder model. Hence, we provided the two endings from the two models and asked the judges to pick a more "sad" ending. The judges picked 32 endings as sad from our model and 18 endings as from the encoder decoder model. The proportions Z test indicates a p-value of 0.03 showing that these results are statistically significant.

\subsubsection{Content relevancy}
%Now that we knew that our model was producing more stylized responses of our choice we wanted to understand the relevancy of these endings. Hence 
To understand the relevancy of the endings produced by our model, we provided the judges with the story context and the two endings from the two models and asked to pick a better story ending. We also gave them the option to pick both the endings if they thought both were relevant to the story context. We observed that the judges picked 23 endings as better from our model, 16 endings from the encoder decoder baseline and 11 as both. This study has a p-value of 0.25 using proportions z test indicating that our model is on par with the encoder decoder model in terms of content relevance.

\subsection{Qualitative Results}
We show some of the example predictions from our model in Table~\ref{table:examples}. In the first example, we see that the model was able to come up with a negative ending and a money minded ending for the same story context based on the adapter that was chosen. For even finer grained styles like peaceful the model could generate outputs that were relevant. In the final example we see that although the model is thinking about property (which is money-minded) the model associates this with the dog in the story context. Hence, we need a way to ground these stories with common sense reasoning as well (perhaps using IE +GA ~\cite{guan2019story}) 
% \rashmicomment{good point, idea for your next paper :P }

\section{Conclusion}
A central tenet of the current language generation models is around controllable generation. While there are decoding algorithms and style transfer techniques to control the generation of text, unfortunately these techniques are not applicable to more finer control in properties like style/persona of text. In this work we show that it is possible to develop generation models that are capable of producing stylized outputs \emph{without} the need of labeled stylized data on the downstream task. Through both automated metrics and human evaluations, we show that our model is better than the Encoder-Decoder baseline by 200\% while maintaining almost same content relevancy. We also show the generalizability of our approach to finer grained styles.

\bibliography{anthology,custom}
\bibliographystyle{acl_natbib}

\appendix

\include{appendix}

\end{document}

%% file: appendix.tex
\section{Appendices}
\label{sec:dendogram}
\begin{figure*}
    \Large
    \vspace{1em}
    \centering
    \includegraphics[width=\textwidth, height=\textheight]{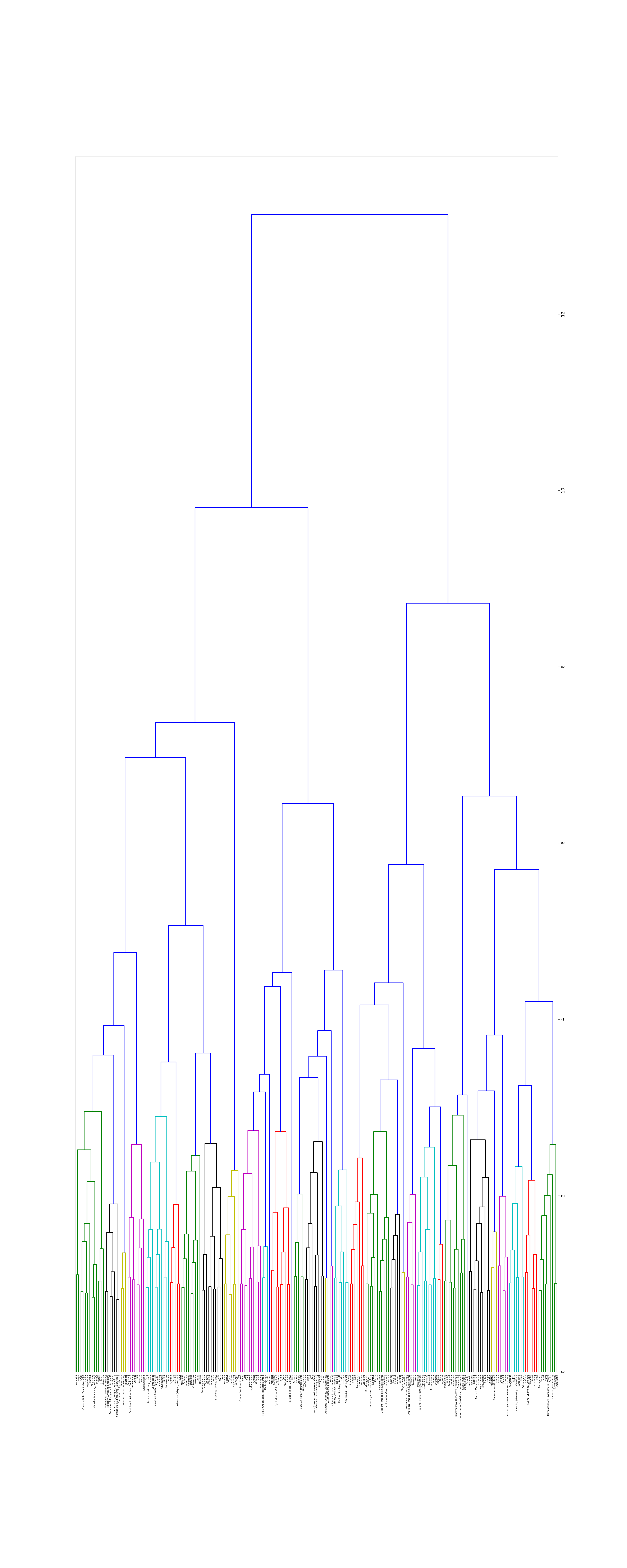}
    \vspace{-5em}
    \caption{Hierarchical clustering of styles using the style embedding matrix generated from the CNN-based text style classifier. Zoom in to view the persona names.}
    \label{fig:my_label}
\end{figure*}

\begin{figure*}[htb!]
\begin{subfigure}{0.50\textwidth}
  \centering
  \includegraphics[width=\linewidth]{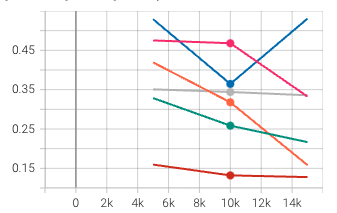}
  \captionof{figure}{Steps VS PIS }
  \label{fig:daily}
\end{subfigure}%
\begin{subfigure}{0.50\textwidth}
  \centering
  \includegraphics[width=\linewidth, height=4cm]{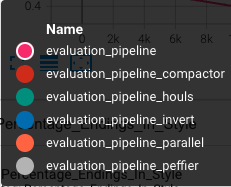}
  \captionof{figure}{Legend}
  \label{fig:mmf}
\end{subfigure}
\caption{We see that for 5/6 adapter types the PIS metric drops as we train the model for longer as the model catastrophically forgets the stylistic aspects and degenerate to an Encoder-Decoder model if you train for too many steps.}
\end{figure*}